# Barriers in Integrating Medical Visual Question Answering into Radiology Workflows: A Scoping Review and Clinicians' Insights

Deepali Mishra [1,2*], Chaklam Silpasuwanchai [1*], Ashutosh Modi[2*] , Madhumita Sushil [3], and Sorayouth Chumnanvej [4]

[1] Asian Institute of Technology, Thailand
[2] Indian Institute of Technology Kanpur (IIT-K), India
[3] University of California, San Fransico
[4] Faculty of Medicine Ramathibodi Hospital, Mahidol University

**ABSTRACT**

**Background**: Medical Visual Question Answering (MedVQA) has emerged as a promising tool for assisting radiologists by automating the interpretation of medical images through question answering. Notwithstanding the significant advancements in MedVQA models and publicly available datasets, the practical integration of MedVQA into clinical settings remains limited.

**Objective:** We aim to systematically examine ongoing research on MedVQA datasets and models, their practical utility, and challenges while considering clinicians' insights.

**Methods:** The study follows the Arksey and O'Malley (2005) scoping review framework. We adopted a two-pronged approach: (1) we systematically examined the study published between September 2018 and September 2024 and mapped the key concepts, advancements, research gap, and challenges in radiology workflow, and (2) we conducted an online survey among experienced clinicians in India and Thailand to gather their perspective on MedVQA's utility and essential features.

**Results:** The scoping review of 68 peer-reviewed studies reveals that nearly 60% of QA pairs are non-diagnostics, offering limited diagnostic value. Most datasets and models lack support for multi-view, multi-resolution imaging, EHRs, and domain knowledge—key components for effective clinical diagnosis. Moreover, there exists a gap between MedVQA evaluation metrics and their clinical relevance.

These concepts are not merely academic. In parallel, the clinician survey, conducted among 50 experienced healthcare professionals (40 from India and 10 from Thailand), confirms this disconnect, as only 29.8% consider current MedVQA systems highly useful in clinical settings. Key clinician concerns include the lack of integration with patient history or domain knowledge (87.2%), preference for manually curated datasets (51.1%), and the need for multi-view image support (78.7%). Additionally, 66% preferred models specialized in single anatomical regions, while a strong 89.4% expressed a preference for dialogue-based interactive MedVQA systems.

[*] Co-corresponding authors:
Deepali Mishra, Ph.D. candidate, Asian Institute of Technology, Thailand. email: st124434@ait.asia
Chaklam Silpasuwanchai, Ph.D., Asian Institute of Technology, Thailand. email: chaklam@ait.asia
Ashutosh Modi, Ph.D., Indian Institute of Technology Kanpur, India. email: ashutoshm@cse.iitk.ac.in

**Conclusion:** Though MedVQA holds promise in radiology workflows, significant challenges, including inadequate multi-modal analysis, insufficient incorporation of patient history, and a disconnect between evaluation metrics and clinical relevance, limit its current impact.

## 1. INTRODUCTION

Approximately 3.6 billion medical imaging examinations, such as X-rays and CT/MRI scans, are performed globally each year[†], underscoring the need for an effective image interpretation system to support clinical decision making. Medical Visual Question Answering (Med VQA), driven by AI, has emerged as a promising healthcare technology to assist interpretation of complex medical images[1–3]. These models aim to streamline diagnosis by answering clinically relevant questions based on medical images. Figure S1 in Supplementary A illustrates a comparative workflow of traditional radiology and the AI-enhanced MedVQA system.

Despite the technical advancements and availability of datasets such as VQA-RAD[4], SLAKE[5], and OVQA[6], Med VQA's real-world clinical utility remains underexplored. Although MedVQA holds substantial promise, many questions addressed in existing datasets offer limited value to radiologists. For example, questions like *"What is the modality of this scan?*[7] *"* with responses such as *"CT*[7] *"* or *"What organ is depicted?*[6] *"* with answers like *"lung*[6]*"* offer minimal assistance in a clinical setting. Ongoing research has predominantly focused on developing models and datasets, often neglecting their utility in clinical practice and clinical integration.

Existing MedVQA surveys[1,2,8] remain fixated on technical performance, sidelining the perspectives of clinicians. This oversight hampers adoption and raises concerns about the practical relevance of current MedVQA systems. Our study addresses this gap by examining how healthcare professionals evaluate the usefulness, trustworthiness, and workflow compatibility of MedVQA in clinical practice.

To address this gap, we adopt a two-pronged approach: (1) a scoping review of existing literature and (2) a targeted survey titled *"Healthcare Professionals' Perspectives on MedVQA Integration."* The survey involved 50 clinicians, 40 from India and 10 from Thailand, including radiologists and other healthcare professionals. While the regional distribution reflects participant availability, we acknowledge the imbalance and interpret findings with caution. By combining insights from literature and clinical stakeholders, this study offers a clinically grounded perspective.

Our review identifies eight core challenges (Figure 1) that hinder the practical adoption of MedVQA systems, based on both technical analysis and clinician feedback. Datasets often lack clinical context, such as patient history (EHRs), external knowledge, and multi-view imaging, while models typically downsample images, losing diagnostic detail. Nearly 60% of QA pairs are non-diagnostic, and current evaluation metrics fail to reflect clinical utility. Most models are adapted from general-domain architectures, lack medical reasoning, and demand high computational resources, limiting their use in low-resource settings. Poor interpretability undermines clinician trust, and integration with systems like PACS or EHRs remains limited. These gaps highlight the urgent need for clinically grounded, efficient, and context-aware MedVQA research, an area largely overlooked in prior reviews.

[†] https://about.cmrad.com/articles/number-of-x-rays-per-year-global-statistics-and-trends

Our main contributions to this paper are as follows:

1. **Radiology-Focused Dataset Analysis:** We examine radiology-specific datasets and assess their clinical relevance.
2. **Question Taxonomy Development:** We propose a taxonomy of questions based on the clinician's perspective and aligned with their clinical utility.
3. **Evaluation Review:** We review the evaluation methods used in peer-reviewed studies on MedVQA models.
4. **Clinician Perspective Integration:** We incorporate insights from a survey conducted with clinicians to understand practical challenges and opportunities.

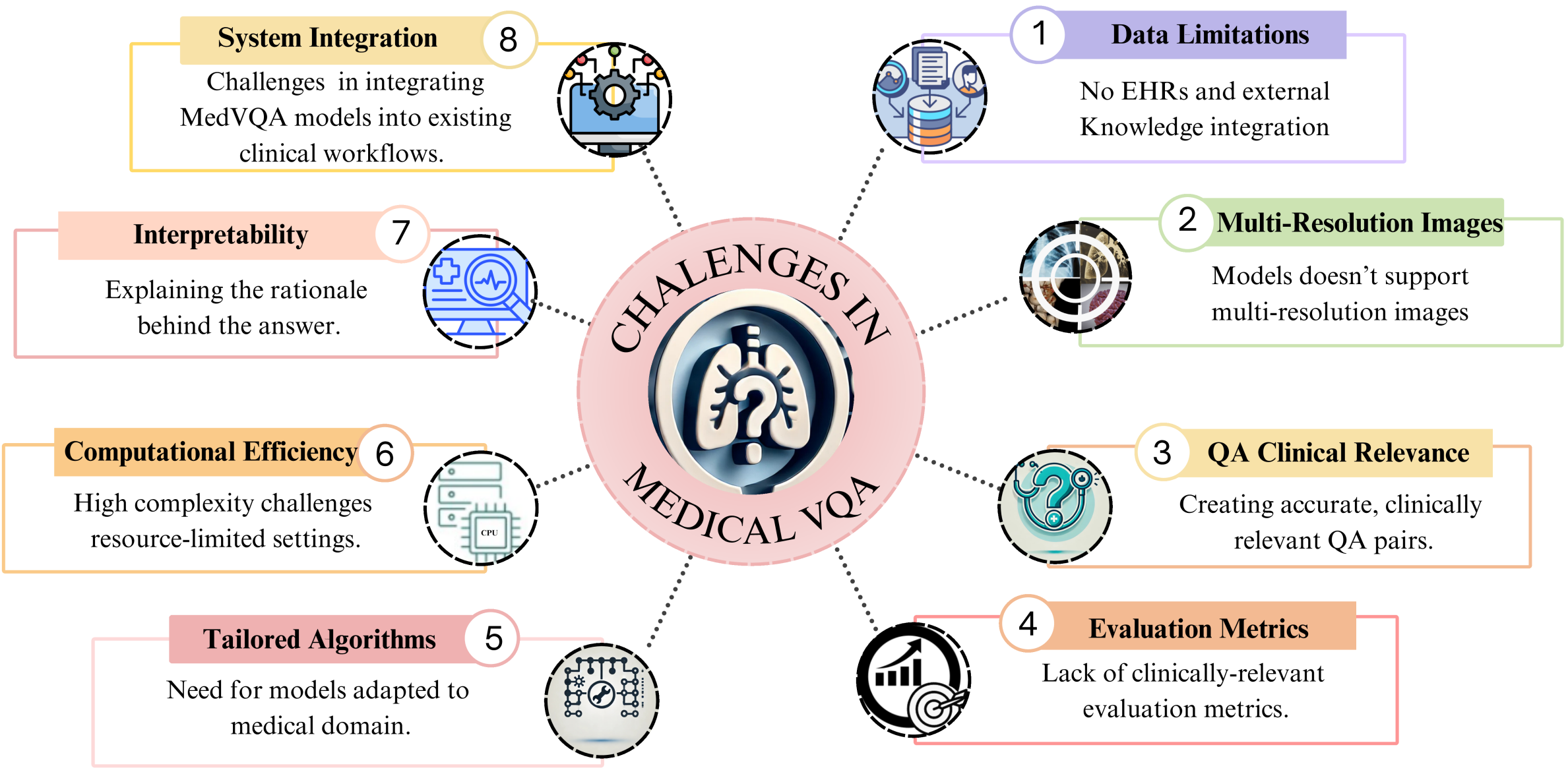

Figure 1: Key Challenges in Integrating Medical VQA in Radiology Workflow.

## 2. METHODOLOGY

### 2.1. Study Design

This scoping review adheres to the Arksey and O'Malley (2005) framework[‡], aiming to map key concepts, and research gaps to assess the extent, scope, and potential for integrating MedVQA systems in clinical settings.

This review aims to address the following research questions:

---

[‡] https://jbi-global-wiki.refined.site/space/MANUAL/355862599/10.1.3+The+scoping+review+framework

**RQ 1:** What are the descriptive statistics of existing MedVQA datasets (e.g., dataset size, supported modalities, curation methods, organs included, question types)?

**RQ 2:** What is the taxonomy and clinical relevance of questions in these datasets?

**RQ 3:** How has MedVQA advanced technically, and what is its clinical utility?

**RQ 4:** How are MedVQA models evaluated? Are these metrics relevant in clinical settings?

**RQ 5:** What are clinicians' insights on the ongoing research on MedVQA systems and their potential impacts on radiology workflows and diagnostic efficiency?

**RQ 6:** What are the future directions for the development of MedVQA, considering clinical relevance?

### 2.2. Screening and Selection of Studies

A comprehensive literature search was performed on Google Scholar using the Python *scholarly*[§] library, targeting peer-reviewed publications from September 2018 to September 2024 to capture developments in MedVQA. Search terms included variations such as "Medical VQA," "Med VQA," "Radiology VQA," "Medical VQA datasets," "VLM & VQA Med," "LLMs & VQA Med," and "Med VQA & ImageCLEF." To ensure quality, only peer-reviewed studies were included; arXiv preprints were excluded due to the lack of formal review. Studies on video-based and surgical VQA were omitted, as they rely on temporal and procedural reasoning distinct from radiology. This review focused exclusively on radiology, the most widely benchmarked domain in MedVQA, supported by established datasets, consistent clinical question types, and direct relevance to diagnostic workflows.

## 3. RESULTS

### 3.1. Overview of Included Studies

A comprehensive search using predefined keywords on Google Scholar yielded 272 studies. After removing the 109 duplicate records, we retained 163 unique studies. Of these 70 studies met the inclusion criteria, out of which 68 were included based on the criteria as shown in Figure 2.

### 3.2. Datasets Used in Medical VQA

A systematic analysis of publicly available MedVQA datasets in radiology reveals key attributes such as image modalities, anatomical coverage, visual question generation (VQG) methods, and external knowledge sources incorporated (Table 1), which is discussed in further detail. Figure 3(a) illustrates the comprehensive analysis of the MedVQA datasets.

---

[§] https://pypi.org/project/scholarly/

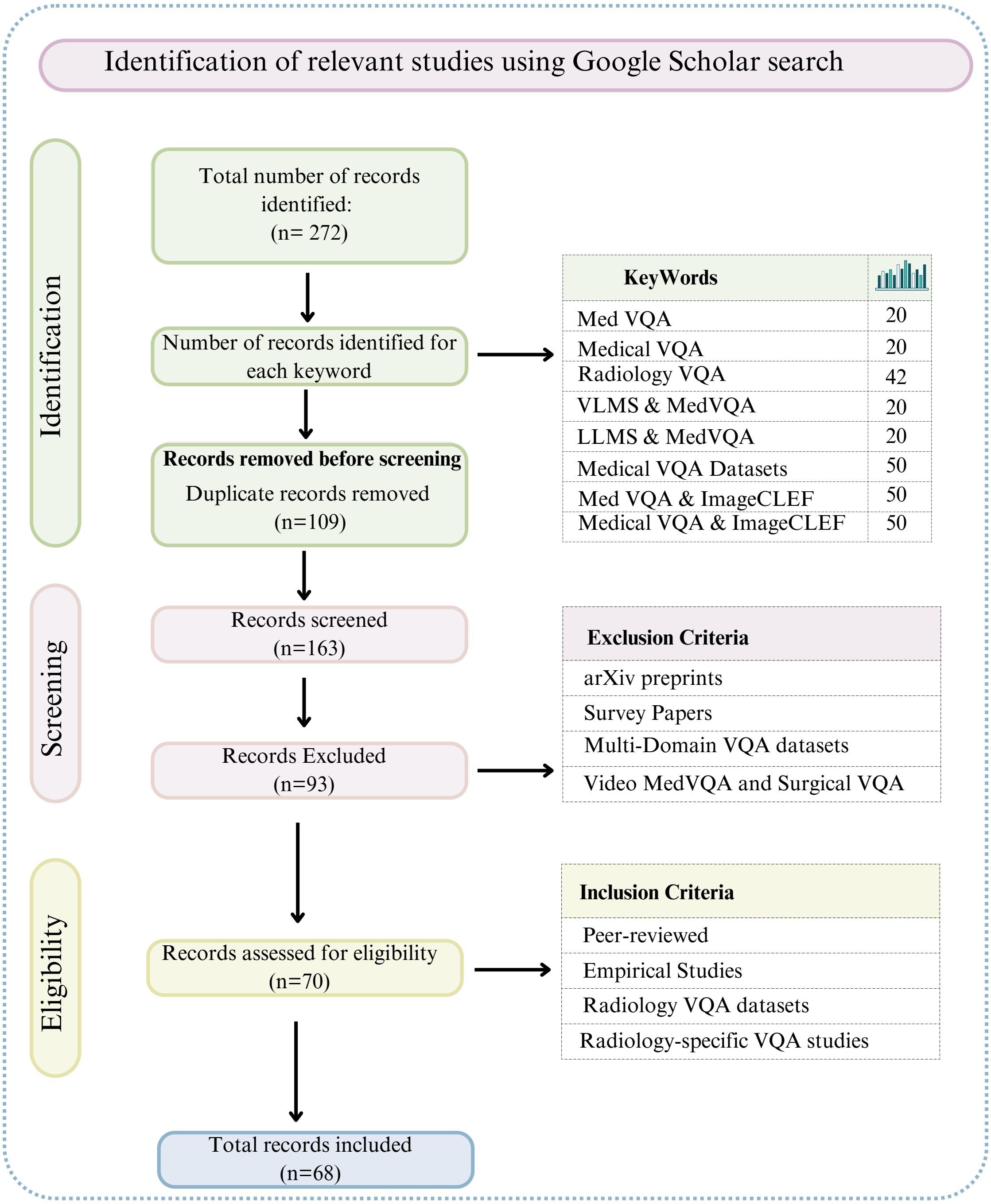

Figure 2: Systematic Literature Retrieval and Evaluation Process for Medical VQA

### 3.2.1: Coverage of anatomical structures across the datasets:

The analysis shows that the chest region is predominantly represented at 24.2 %, as evidenced by datasets such as VQA Med 2018[7], RadVisDial[10], Medical Diff VQA[11], and MIMIC CXR VQA[9]. Head (18.1%) and abdomen (15.1%) follow next, while other regions are covered at much lower rates (ranging from 3.0% to 6.0%).

This imbalance in anatomical representation, across datasets like VQA-RAD[4], SLAKE[5], OVQA[6], VQA Med 2018[7], and VQA Med 2019[12], highlights a critical disparity that could impact their applicability and generalizability in real-world clinical scenarios.

### 3.2.2: Dataset Curation Method

The methodology for curating Visual Question Generation (VQG) significantly influences the quality and size of MedVQA datasets. Three primary approaches are utilized: manual, semi-automatic, and automatic.

**Manual (27.3%):** In this method, clinicians create question-answer pairs based on images, ensuring accuracy but limiting dataset size, examples: SLAKE[5], VQA RAD[4], and RadVisDial (Gold standard)[10].

**Semi-Automatic (54.5%):** This approach uses automated generation from image captions from multiple sources, like MIMIC CXR[**], PubMed articles[††], and the MedPix database[‡‡], with clinicians' validation. It balances accuracy and data size, as demonstrated by OVQA[6], VQA-Med-2018[7], VQA-Med-2019, VQA-Med-2020[13], VQA-Med-2021[14], and Medical-Diff-VQA[11].

**Automatic (18.2%)**: Automatic question-answer pairs are generated from image captions or labels, enabling large datasets but introducing noise, making them inaccurate for clinical settings —for example, RadVisDial (Silver Standard)[10].

### 3.2.3 Modality and Anatomical Structures Covered.

**Single-anatomy and single-modality:** Around 30% focus solely on the chest anatomy, highlighting their value in answering questions related to chest abnormalities. Datasets include MIMIC CXR VQA[9], Medical-Diff-VQA[11], and RadVisDial (Gold and Silver Standard)[10].

**Multi-anatomy and multi-modality:** Approximately 60% of datasets incorporate multiple imaging modalities, such as X-rays, CT scans, MRIs, and ultrasounds, covering anatomical structures such as the head, chest, abdomen, pelvis, etc. Example datasets include VQA RAD[4], SLAKE[5], OVQA[6], VQA-Med 2018[7], VQA-Med 2019[12], and VQA-Med 2020[13]. These datasets are more suitable for generalization across diverse tasks.

**3.2.4: External Knowledge Integration:** Only 20% of the datasets incorporate external knowledge, although clinicians routinely make diagnoses based on their experience, patient history, and medical imaging. The SLAKE[5] dataset integrates external knowledge through a knowledge graph, and MIMIC-CXR VQA[9] incorporates electronic health records (EHRs), enhancing their applicability in real-world settings.

---

[**] https://physionet.org/content/mimic-cxr

[††] https://pubmed.ncbi.nlm.nih.gov/

[‡‡] https://medpix.nlm.nih.gov/

Table 1: Descriptive Analysis of Existing MedVQA Datasets

| Modality | Datasets/ year | Best Models | No. of images | No. of QA pair | Source of Images | QA Creation Methods | Anatomy Covered | Question Type |
|---|---|---|---|---|---|---|---|---|
| X-ray | RadVisDial (2019) (Silver Standard)[10] | SAN VQA[10] | 91,060 | 455,300 | MIMIC-CXR | Automatic | Chest | Abnormality |
| | RadVisDial (2019) (Gold Standard) [10] | SAN VQA[10] | 100 | 500 | | Manually | Chest | Abnormality |
| | Medical-Diff-VQA(2023)[11] | MMQ[15] | 16,432 4 | 700,703 | MIMIC-CXR | Semi-Automatic | Chest | Abnormality, Presence, View, Location, Level, Type, Difference |
| | MIMIC CXR VQA (2023) [9] | M3AE[16] | 142,79 7 | 377,391 | MIMIC-CXR, Chest-ImaGenome | Automatic | Chest | Anatomy, Attribute, Presence, Abnormality, Plane, Gender, Size |
| X-ray/ CT scan/ MRI/ Ultrasound | VQA-Med-2018 [7] | Chakri[17] | 2,886 | 6,413 | PubMed | Semi- | Abdomen, Head, other | Location, Finding, Yes/No |
| | VQA-RAD- 2018[4] | Q2ATransformer[18] | 315 | 3,515 | MedPix Database | Manually | Head, Chest, Abdomen | Modality, Plane, Color, Size, Attribute, Counting, Organ system, Abnormality, Presence |
| | VQA-Med 2019[12] | MedfuseNet[19] | 4,200 | 15,292 | MedPix Database | Semi-Automatic | Breast, skull and Face, sinuses, neck, Spine, Musculo-skeletal, Heart, Lungs, Gastrointestinal, Vascular Genitourinary, lymphatic | Modality, Plane, Organ system, Abnormality |
| | VQA-Med 2020[13] | AIML[20] | 5,000 | 5,000 | MedPix Database | Semi-Automatic | | Abnormality |
| | SLAKE (2021) [5] | CMMO[21] | 642 | 14,000 | MedPix Database | Manually | Neck, Chest, Abdomen, Pelvic, Head | Organ, Location, Appearance, Shape, Position, Knowledge graph, Abnormality, Modality, Plane, Quality, Color, Size, Shape |
| | VQA-Med 2021[14] | SYSU_HCP[22] | 5,000 | 5,000 | MedPix Database | Semi-Automatic | | Abnormality |
| | OVQA (2022)[6] | MMBERT[23] | 2,001 | 19,020 | Electronic Medical Record | Semi-Automatic | Chest, Head, Leg, Hand | Abnormality, Attribute other, Condition Presence, Modality, Organ System, Plane |

**3.2.5: Dialogue-Based Datasets:** RadVisDial (Gold and Silver Standard)[10] is the only dataset, accounting for 10% of the total, with structured questions in a dialogue format, simulating real-world clinician-patient interactions.

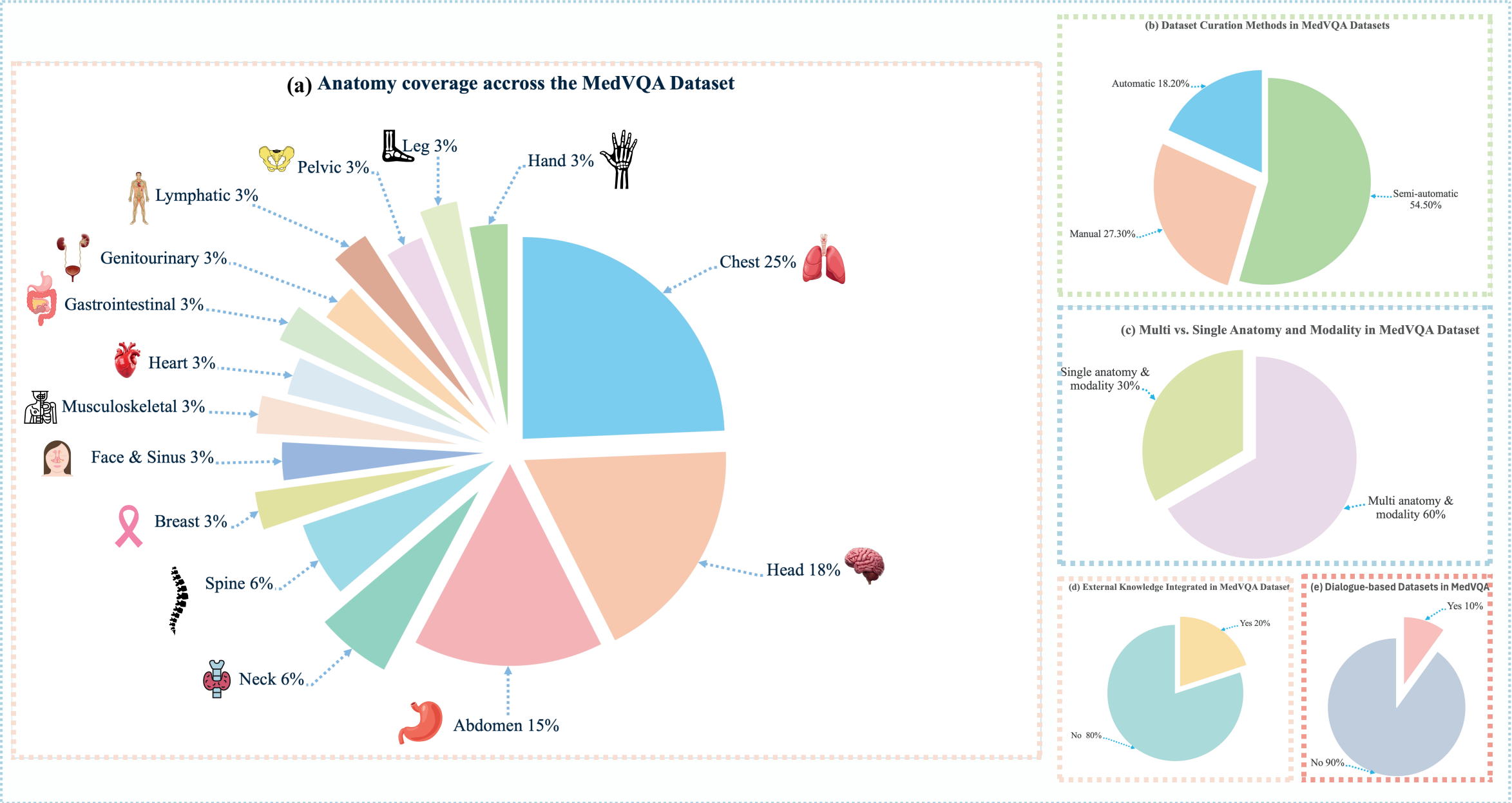

(a): Comprehensive Overview of Medical VQA Datasets: Anatomical Coverage, Curation Methods, Modality & Anatomy Distribution, and Knowledge Integration

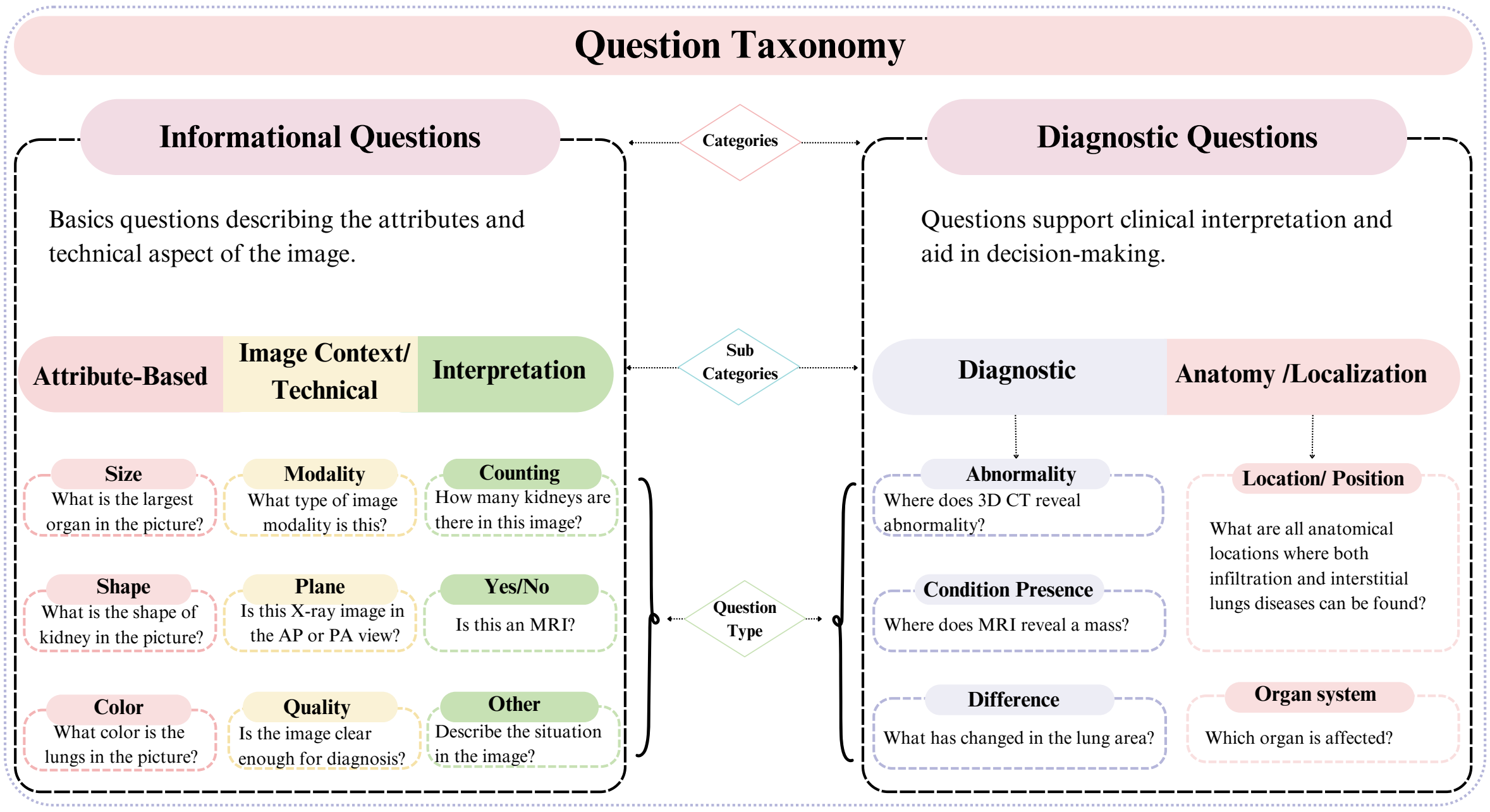

(b) Taxonomy of Questions for Medical Visual Question Answering (Med-VQA)

Figure 3: A Comprehensive Analysis of the MedVQA Dataset. (a) Illustrates the various attributes of existing MedVQA Datasets. (b) Illustrate the taxonomy of the questions introduced.

### 3.3 Taxonomy and Clinical Relevance of Questions

To assess the clinical relevance of questions in existing MedVQA datasets, we developed a structured, manually developed taxonomy. We analyzed existing datasets to identify question types they address, as shown in Table S1(a) in the Supplementary B. Our taxonomy follows the principles of mutual exclusivity. We categorized questions into non-diagnostic and diagnostic questions, based on clinical relevance from the clinician's perspective, illustrated in Figure 3 (b)

### 3.3.1 Non-Diagnostic Questions

The non-diagnostic questions are attribute-based questions (e.g., size, shape, and color of anatomical structures), technical questions (e.g., modality, plane, quality), and basic interpretation questions (e.g., closed-ended questions, counting). While these questions are important for the model to understand the characteristics of images, they do not contribute directly to a diagnosis.

### 3.3.2 Diagnostic Questions

Diagnostic questions directly support clinical decision-making by enhancing the understanding of medical findings. These include abnormality detection, condition presence, anatomical localization (e.g., affected region or organ), and clinical interpretation.

## 3.4. Advancements in MedVQA Methodologies for Radiology Datasets

The first Medical Visual Question Answering (MedVQA) was introduced in ImageCLEF 2018[§§]. Since then, the field has continuously evolved through advancements in deep learning architectures from Convolutional Neural Networks (CNN)[17,24–38], Residual Networks (ResNet)[19,22,23,39–50], Long Short-Term Memory (LSTM)[15,19,39–41,49–53], Vision Transformers (ViTs)[23,28,34,36,44,45,47,54], BERT models[19,23,28,33,34,36–38,44,45,47,54], etc. The evolution of multimodal learning expanded with Vision Language Models (VLMs)[55–60] and Multimodal Large Language Models (MLLMs)[55,59,61]. Additionally, techniques such as self-supervised Learning[23,62–64], meta-learning[15,16,51,65], Contrastive Learning[56,58,59,64] , and parameter-efficient tuning[66] have significantly enhanced MedVQA models, improving their adaptability and generalization.
Despite continuous developments to create a robust model, several limitations persist, impacting its clinical relevance and effectiveness in real-world clinical settings. Supplementary B Table S1 (b) illustrates the advancement of the deep learning model across various datasets.

**Image Resolution Constraints:** Most Med VQA models take input images of 224x224 resolution. This, in turn, results in data loss, particularly in high-resolution images such as CT and MRIs, where the fine-grained semantic features are crucial for accurate diagnosis.

**Multi-view Imaging:** As stated by the American College of Radiology (ACR)[***], a minimum of two views is required for accurate diagnosis in imaging, such as X-rays (e.g., Anterior and Posterior opinions). Multiple views across different planes are essential in CT and MRI to ensure a comprehensive and accurate diagnosis. However, none of the reviewed models support multi-view image input, inadvertently limiting their ability to provide a thorough diagnostic assessment, thereby reducing their clinical relevance.

[§§] https://www.imageclef.org/2018
[***] https://www.acr.org/Clinical-Resources/Clinical-Tools-and-Reference/Appropriateness-Criteria

**Task Formulation Limitation:** Approximately 85% of studies treat the MedVQA task as classification, selecting answers from a predefined set. However, free-text generation models like MMBERT[23], MISS[63], BiomedCLIP[59], PubMedCLIP[56], LlaVA Med[55], and Med Flamingo[60] offer greater clinical relevance by providing nuanced answers. In Supplementary Table S1(b), models are categorized based on their approach: classification**,** generation**,** or a combination of both, while Table 1 highlights the best-performing models across various datasets.

**Pretraining Bottlenecks:** Approximately 87% of MedVQA models are pre-trained on general image-text datasets rather than medical-specific corpora. Only a few, such as PubMedCLIP[56], CMMO[21], BioMedCLIP[59], Med-Flamigo[60], and PMC-CLIP[57], incorporate domain-specific pretraining, which is crucial for giving context-aware answers.

**Evaluation Metrics and Clinical Interpretability:** Most models use metrics like BLEU score and accuracy to validate the output, but they do not adequately reflect the clinical interpretability and practical utility of MedVQA.

### 3.5. Evaluation of Clinical Relevance

The MedVQA task is treated as both a classification and a generation task. In classification tasks, where a single correct answer is expected, metrics such as accuracy[2], recall[2], precision[2], and F1 score[2] are used to assess the correctness of the answer. In contrast, generation tasks involve free-text and are evaluated using metrics like BiLingual Evaluation Understudy (BLEU)[2], Concept-based Semantic Similarity (CBSS)[2], Word-based Semantic Similarity (WBSS)[2], and Metric for Evaluation of Translation with Explicit Ordering (METEOR)[2] are used to assess the accuracy and relevance of the generated responses. These metrics fail to assess the clinical relevance of the models, such as interpretability and ease of integration into clinical workflows. A comprehensive evaluation framework is required to assess accuracy and real-world clinical applicability.

### 3.6. Feedback from Healthcare Professionals:

An online survey titled "Healthcare Professionals' Perspectives on the Integration of Medical Visual Question Answering (MedVQA)" was conducted using Google Forms among 50 clinicians from India (n = 40) and Thailand (n = 10), and approved by the Institutional Review Board (IRB) Reference No. RERC 2024/047. The survey was necessary to assess whether current MedVQA developments align with clinical priorities and to identify gaps that may not be evident through literature alone.

The survey featured multiple-choice questions on MedVQA dataset characteristics and improvement areas (see Supplementary C). Respondents were surgeons (46.8%), radiologists (31.9%), and physicians (21.3%), with 74.5% having over 10 years of experience. Despite regional imbalance, consistent cross-country responses support the reliability of the findings. Clinician feedback is summarized in Figure 4.

**Limited AI awareness**: Only 8.5% of clinicians use AI tools regularly, whereas 91.5% are not using AI regularly or are not even aware of it.

**MedVQA Utility:** Only 29.8% reported it highly beneficial in clinical workflow, while 57.4% found it moderately helpful. Clinicians highlighted diagnostic recommendations as the most important feature (21.7%), followed by image interpretation (17.4%) and integration with radiology systems such as PACS (15.2%). Additionally, 51.1% preferred dialogue-based interaction over single-turn QA.

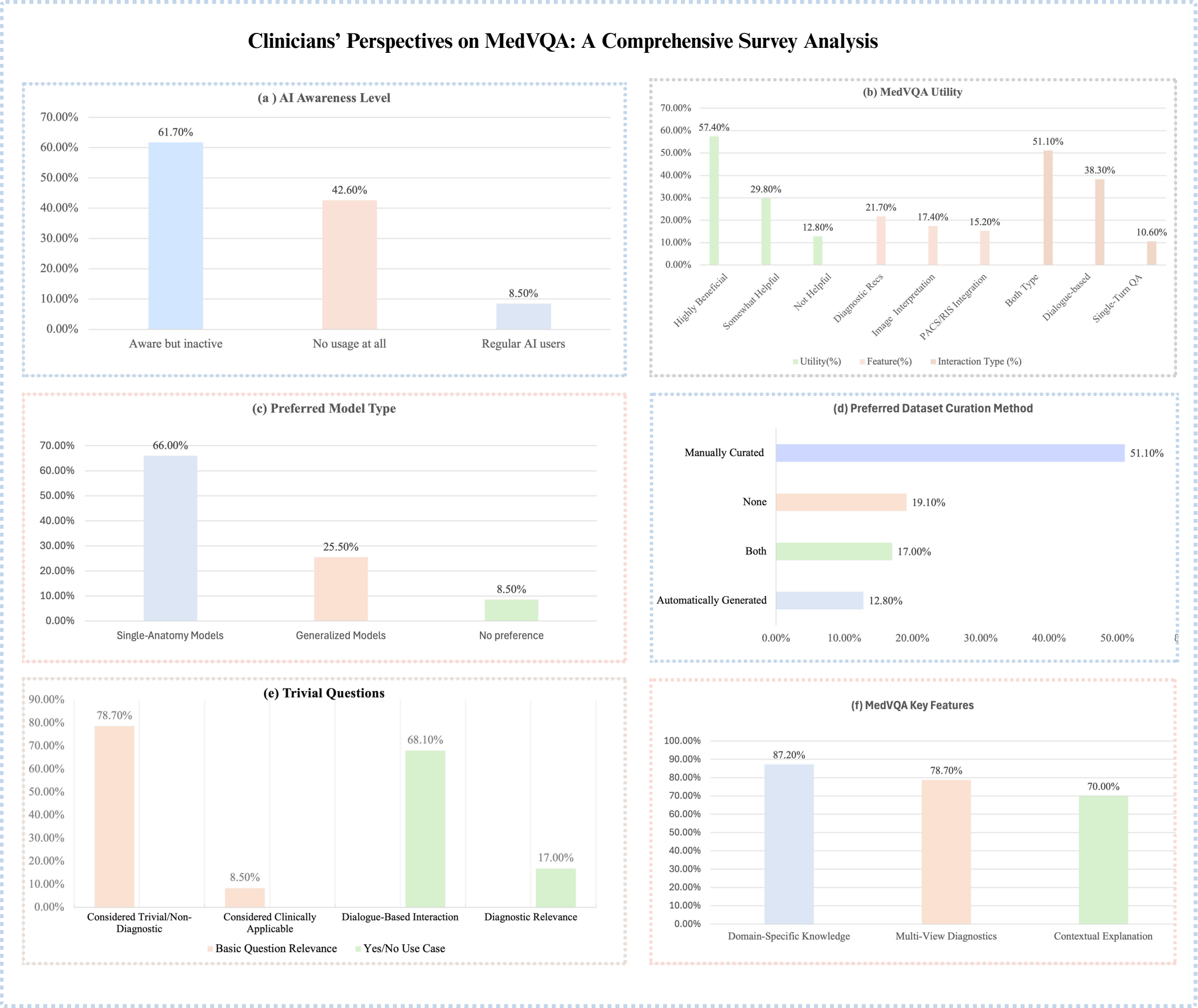

Figure 4: Clinicians' Perspectives on MedVQA: A Comprehensive Survey Analysis

**Anatomy-Specific Models:** Around 66% of clinicians prefer models focused on a single anatomical region; however, only 25.5% prefer models that can be generalized across various anatomical regions.

**Dataset:** Around 51.1% of clinicians preferred manually curated datasets, while automatically generated datasets had limited acceptance, with only 12.8%.

**Trivial Questions:** Basic questions about modality, size, color, or organ presence were considered descriptive and trivial by 78.7% of clinicians, and only 8.5% believe it is clinically applicable. Close-ended questions that give answers as yes/no were more suitable for dialogue-based interaction, with 68.1%, followed by 17% considering it to be diagnostic relevant.

**MedVQA Key Features:** Clinicians emphasized key MedVQA features, with 70% valuing contextual explanation for the answers. 78.7% preferred models capable of multi-view diagnostics. Incorporating domain-specific knowledge was important to 87.2% of clinicians.

## 4. DISCUSSION

This discussion synthesizes the core challenges in current MedVQA research and future directions, integrating both technical insights and clinician perspectives. While much of our analysis centers on datasets and model design, the clinician survey also highlighted broader issues, such as interpretability, deployment efficiency, and clinical integration, that are critical for real-world adoption. Addressing these limitations is essential for bridging the translational gap.

### 4.1. Dataset Limitations

Most MedVQA datasets are limited to single-turn image-based QA pairs and lack integration with electronic health records (EHRs), structured diagnostic reports, and multi-view (e.g., anterior and posterior) inputs. Future datasets should incorporate EHR-linked imaging data, standardized medical ontologies, diagnostic context, and support for multi-view image inputs. QA pairs should be curated or validated by radiologists, ideally using semi-automated methods, such as LLM-generated QA pairs followed by radiologist validation.

### 4.2. Inadequate Handling of Multi-Resolution Data

Although MedVQA datasets include multi-resolution images such as X-ray, CT, and MRI, most models downsample inputs to a fixed resolution, causing loss of crucial semantic information. Future models should incorporate resolution-aware mechanisms, such as dynamic rescaling or patch-based attention, to preserve important details. Additionally, adopting modality-specific architectural components, such as separate processing branches or embeddings for each imaging modality, may improve representation quality and model generalizability.

### 4.3. Limited Clinical Relevance of QA Pairs

Approximately 60% of QA pairs in the current datasets are non-diagnostic and limited to single-turn interactions. To improve clinical utility, future datasets should adopt a dialogue-based format that includes questions related to abnormalities, differential diagnosis, and treatment planning. One effective strategy is to extract QA pairs from structured radiology reports using natural language processing. Additional diversity can be introduced by simulating clinical scenarios such as follow-ups, uncertainty-driven queries, or treatment-related decisions. Models should also incorporate session memory or a context window to handle follow-up or clarification questions.

### 4.4. Misaligned Evaluation Metrics

Conventional NLP metrics such as BLEU, accuracy, and F1-score are inadequate for evaluating MedVQA systems in clinical contexts, as they emphasize surface-level similarity rather than diagnostic correctness. Clinician feedback highlights the need for evaluation metrics that reflect semantic understanding, clinical validity, and patient safety. Domain-specific tools developed for radiology report evaluation, such as CheXbert[67] and RadGraph[68] ,

which assess entity- and relation-level accuracy, offer promising foundations for designing more clinically relevant MedVQA benchmarks.

### 4.5. Lack of Medical-Specific Modeling

Most MedVQA models are adapted from general-purpose vision-language architectures that are not well-suited to the specialized terminology and structured reasoning required in medicine. Future models should incorporate domain-adaptive pretraining, retrieval-augmented generation with clinical knowledge bases. Techniques such as graph-based reasoning and structured attention informed by ontologies can further improve alignment with clinical reasoning tasks.

### 4.6. Computational Challenges

Transformer-based MedVQA models are resource-intensive, hindering real-time use in clinical settings, particularly in low-resource environments. To enable broader deployment, future research should explore efficiency-boosting techniques such as pruning, quantization, mixture of experts (MoE), and knowledge distillation to reduce computational overhead.

### 4.7. Lack of Interpretability

Clinicians reported low trust in MedVQA models as they lack interpretability, providing answers without explaining how they were derived or highlighting the visual cues that informed the response. This absence of reasoning, traceability, and visual justification undermines. For clinical adoption, it is essential to develop models that offer interpretable outputs, such as attention maps, answer rationales, and confidence scores, that clearly indicate how and why a particular answer was generated.

### 4.8. Poor System Integration

Despite advances in modeling, most MedVQA systems remain disconnected from clinical systems like PACS or EHR platforms. This lack of interoperability, particularly with standards like DICOM, hinders practical deployment within existing clinical workflows. Future research should focus on system-level integration, enabling MedVQA models to operate as plug-and-play modules within existing healthcare IT systems.

## 5. CONCLUSION

This paper highlights the persistent gap between technical progress in MedVQA and its clinical applicability. Through a scoping review and clinician survey, we identify eight critical challenges limiting real-world deployment: inadequate dataset diversity and context, poor handling of multi-resolution inputs, limited clinical relevance of QA pairs, misaligned evaluation metrics, lack of domain-specific modeling, computational inefficiency, low interpretability, and insufficient integration with clinical systems. Clinicians emphasized the need for features such as dialogue-based QA, multi-view imaging, external knowledge integration, curated datasets, and contextual reasoning. Addressing these challenges through clinically grounded datasets, efficient and interpretable models, and system-level interoperability will be key to realizing MedVQA's potential to improve diagnostic accuracy, decision-making, and workflow efficiency in real-world medical settings.

## SUPPLEMENTARY

### Supplementary A:

Figure S1 illustrates a comparative diagnostic workflow, highlighting the traditional radiology process and the enhanced workflow with MedVQA integration.

### Figure S1 (a): Traditional Diagnostic Workflow

Traditionally, a patient visits the hospital for a consultation with a doctor. If deemed necessary, radiology imaging is then conducted. Once the imaging is completed, a radiologist reviews the images and prepares a diagnostic report.

### Figure S1(b): Diagnostic Workflow with MedVQA System

By integrating AI-powered support, the MedVQA system speeds up the diagnostic process in radiology. The workflow starts with the patient visits. Upon consultation, radiology imaging is suggested if it is deemed necessary. After the imaging, the scanned system processes the MedVQA image, and the corresponding questions are answered. This technology improves total patient care by increasing diagnostic accuracy, decreasing burden, speeding up decision-making, and integrating seamlessly with current workflows.

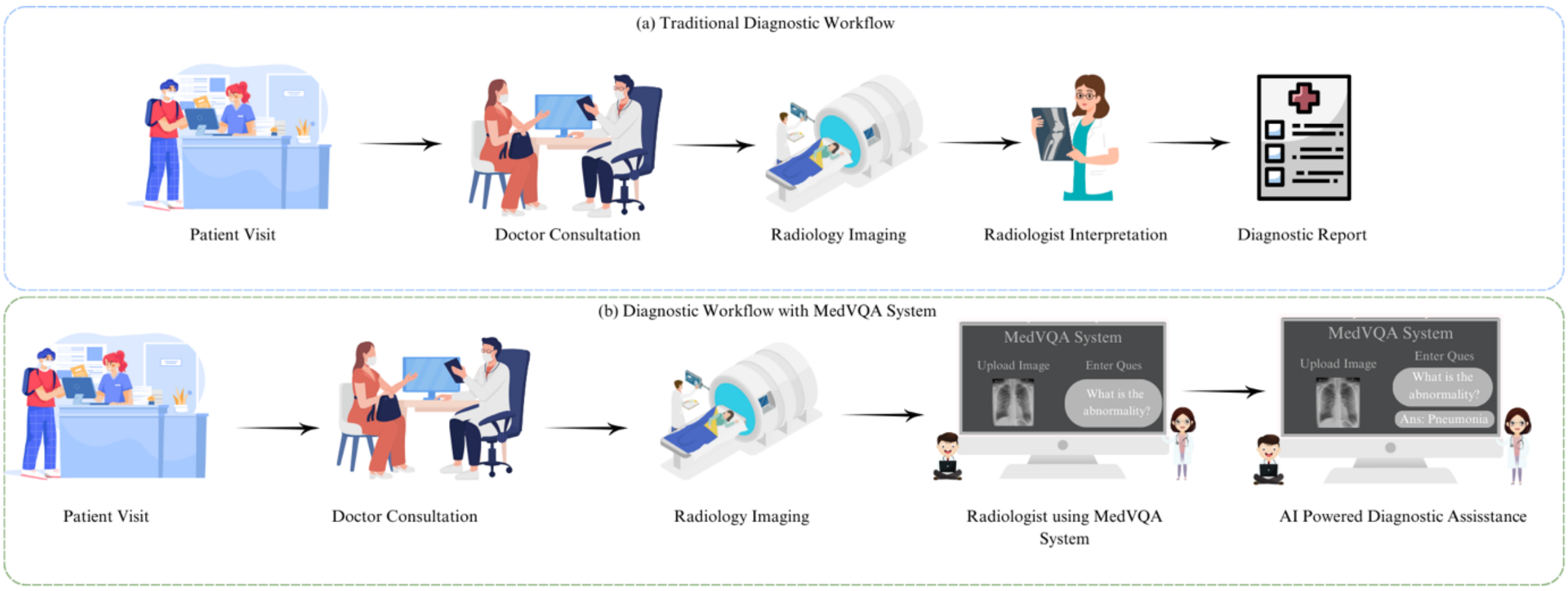

Figure S1: Comparative Diagnostic Workflow: Traditional vs. MedVQA-Assisted Approach

### Supplementary B

### Table S1(a) Overview of Taxonomy Questions Across Various Datasets

| Question Taxonomy | Question Taxonomy | Content-Type | Question-Answer Samples |
|---|---|---|---|
| **VQA Med 2018** | | | |
| **Technical Questions** | Informational Questions | Modality | What type of image modality is this? |
| **Anatomical Questions** | | Organ | Where does an MRI reveal a mass? |
| **Attribute Type** | | Plane | Where does MRI show the parotid gland mass? |
| **Diagnostic Questions** | Diagnostic Questions | Abnormality | Where does 3D CT reveal abnormality? |
| **VQA RAD** | | | |
| **Technical Questions** | Informational Questions | Modality | Is this an MRI? |
| **Anatomical Questions** | | Organ | What is the organ system? |
| | | Presence | Is there gastric fullness? |
| **Attribute Type** | | Plane | Is this an axial image? |
| | | Size | What is dilated? Ans: duodenum |
| | | Colour | Is the lesion more or less dense than the liver? |
| | | Other | Is the mass well circumscribed? Ans:No |
| **Diagnostic Questions** | Diagnostic Questions | Position | What is the location of the mass? |
| | | Abnormality | Which organ is affected? |
| **VQA Med 2019** | | | |
| **Technical Questions** | Informational Questions | Modality | Is this a t1-weighted, t2-weighted, or flare images? |
| **Anatomical Questions** | | Organ | What organ system is shown in this x-ray? |
| **Attribute Type** | | Plane | What is the plane of this MRI? |
| **Diagnostic Questions** | Informational Questions | Abnormality | What abnormality is seen in the image? |
| **VQA Med 2020** | | | |
| **Diagnostic Questions** | Diagnostic Questions | Abnormality | what is abnormal in the CT scan? |
| **VQA Med 2021** | | | |
| **Diagnostic Questions** | Diagnostic Questions | Abnormality | What is abnormal in the x-ray? |
| **RadVisDial** | | | |
| **Diagnostic Questions** | Diagnostic Questions | Abnormality | **Dialogue:**<br>Q: Airspace opacity?<br>Q: Fracture?<br>Q: Lung lesions? |
| **OVQA** | | | |
| **Technical Questions** | Informational Questions | Modality | What imaging modality is seen here? |
| **Anatomical Questions** | | Organ | What organ system is pictured? |
| | | Presence | Are there any fractures? |
| **Attribute Type** | | Plane | Is this a sagittal plane? |
| | | Other | Describe the situation in the image? |
| **Diagnostic Questions** | Diagnostic Questions | Abnormality | What abnormalities are seen within the ulna? |
| | | **SLAKE** | |
| **Technical Questions** | Informational Questions | Modality | What modality is used to take this image? |
| **Question Types** | | Close-ended | Is the chest x-ray normal? |
| | | Open-ended | What is the scanning plane of this image? |
| **Attribute Type** | | Plane | What is the scanning plane of this image? |
| | | Size | What is the largest organ in the picture? |
| | | Shape | What is the shape of kidney in the picture? |
| | | Color | What color is the lungs in the picture? |
| | | Other | How many kidneys are there in this image? |
| **Anatomical Questions** | | Organ | Does the picture contain liver? |
| **Diagnostic Questions** | Diagnostic Questions | Position | Which part of the body does this image belong to? |
| | | Abnormality | Is the lungs healthy? |
| | | Extra knowledge | What is the effect of the main organ in this picture? |
| | | **MIMIC CXR VQA** | |
| **Anatomical Questions** | Diagnostic Questions | Presence | Does the cardiac silhouette show any evidence of diseases or devices? |
| | | Position | What are all anatomical locations where both infiltration and interstitial lungs diseases can be found? |
| **Diagnostic Questions** | | Abnormality | Are there signs of abnormalities in both the left lungs and the right lungs? |
| | | Extra knowledge | Please specify the patient's gender. |
| **Attribute Type** | Informational Questions | Plane | Is this X-ray image in the AP or PA view? |
| | | Size | Is the cardiac silhouette's width larger than half of the total thorax width? |
| | | Other | List all detected anatomical findings. |
| | | **Med-Diff- VQA** | |
| **Anatomical Questions** | Diagnostic Questions | Presence | Is there any <abnormality>? |
| | | Position | Where in the image is the <abnormality> located? |
| **Diagnostic Questions** | | Abnormality | What abnormalities are seen in the image? |
| | | Difference Questions | What has changed in the <location> area? |
| **Attribute Type** | Informational | Plane | Which view this image taken? |
| | Questions | Other | What type of <abnormality>? |

**Table S1 (b): Advancements of MedVQA models across various datasets.**

$\mathcal{M}_\mathcal{I}$: Image encoder pre-trained on medical images, $\mathcal{N}_\mathcal{I}$: Image encoder pre-trained on natural images, $\mathcal{M}_\mathcal{T}$: Text encoder pre-trained on bio-medical images, $\mathcal{N}_\mathcal{T}$: Text encoder pre-trained on general text,$\mathcal{N}/\mathcal{A}$: Not mentioned. $\mathcal{G}$ : the MedVQA task is treated as generation, $\mathcal{C}$: the MedVQA task is treated as classification, and $\mathcal{B}$ : the MedVQA task is treated as both.

| Model | Max resolution | Image Encoder Pre-training | Language Encoder Pre-training | Type of Task | BLEU | Accur-acy |
|---|---|---|---|---|---|---|
| **VQA-Med-2018** | | | | | | |
| Chakri[17] | 224x224 | $\mathcal{N}_\mathcal{I}$ | $\mathcal{N}/\mathcal{A}$ | $\mathcal{G}$ | 0.188 | |
| BPI-MVQA[69] | 224x224 | $\mathcal{N}_\mathcal{I}$ | $\mathcal{M}_\mathcal{T}$ | $\mathcal{B}$ | 0.168 | |
| UMMS[40] | - | $\mathcal{N}_\mathcal{I}$ | $\mathcal{N}_\mathcal{T}$ | $\mathcal{C}$ | 0.162 | |
| HQS-VQA[42] | 224x224 | $\mathcal{N}_\mathcal{I}$ | $\mathcal{N}_\mathcal{T}$ | $\mathcal{C}$ | 0.132 | |
| NLM[25] | | $\mathcal{N}_\mathcal{I}$ | $\mathcal{N}_\mathcal{T}$ | $\mathcal{C}$ | 0.121 | |
| JUST[26] | - | $\mathcal{N}_\mathcal{I}$ | $\mathcal{N}/\mathcal{A}$ | $\mathcal{C}$ | 0.061 | |
| **VQA RAD 2018** | | | | | | |
| Q2ATransformer[18] | 224x224 | $\mathcal{N}_\mathcal{I}$ | $\mathcal{N}_\mathcal{T}$ | $\mathcal{B}$ | | 0.804 |
| CMMO[70] | 384x384 | $\mathcal{M}_\mathcal{I}$ | $\mathcal{M}_\mathcal{T}$ | $\mathcal{G}$ | | 0.796 |
| M2I2[62] | 384x384 | $\mathcal{N}_\mathcal{I}$ | $\mathcal{N}_\mathcal{T}$ | $\mathcal{G}$ | | 0.768 |
| MISS[63] | 480x480 | $\mathcal{N}_\mathcal{I}$ | $\mathcal{N}_\mathcal{T}$ | $\mathcal{G}$ | | 0.760 |
| BiRL[71] | 128x128 | $\mathcal{N}/\mathcal{A}$ | $\mathcal{N}_\mathcal{T}$ | $\mathcal{C}$ | | 0.760 |
| VQA-Adapter[66] | 224x224 | $\mathcal{N}_\mathcal{I}$ | $\mathcal{N}_\mathcal{T}$ | $\mathcal{C}$ | | 0.758 |
| VL [72] | 224x224 | $\mathcal{N}_\mathcal{I}$ | $\mathcal{N}_\mathcal{I}$ | $\mathcal{B}$ | | 0.754 |
| MHKD-MVQA[73] | - | $\mathcal{N}_\mathcal{I}$ | $\mathcal{N}_\mathcal{T}$ | $\mathcal{G}$ | | 0.736 |
| WADAN (NP)[74] | - | $\mathcal{N}_\mathcal{I}$ | $\mathcal{N}_\mathcal{T}$ | $\mathcal{B}$ | | 0.735 |
| CPRD[50] | 224x224 | $\mathcal{N}_\mathcal{I}$ | $\mathcal{N}_\mathcal{T}$ | $\mathcal{C}$ | | 0.727 |
| BPI-MVQA[69] | 224x224 | $\mathcal{N}_\mathcal{I}$ | $\mathcal{M}_\mathcal{T}$ | $\mathcal{B}$ | | 0.727 |
| MITER[64] | 448x448 | $\mathcal{M}_\mathcal{I}$ | $\mathcal{M}_\mathcal{T}$ | $\mathcal{C}$ | | 0.721 |
| MMBERT(Gen.) [23] | 224x224 | $\mathcal{N}_\mathcal{I}$ | $\mathcal{N}_\mathcal{T}$ | $\mathcal{G}$ | | 0.720 |
| QCR[46] | 255x255 | $\mathcal{N}/\mathcal{A}$ | $\mathcal{N}/\mathcal{A}$ | $\mathcal{C}$ | | 0.716 |
| VB-MVQA[74] | 224x224 | $\mathcal{N}_\mathcal{I}$ | $\mathcal{N}_\mathcal{I}$ | $\mathcal{B}$ | | 0.713 |
| CopVQA | 224x224 | $\mathcal{N}_\mathcal{I}$ | $\mathcal{N}_\mathcal{T}$ | $\mathcal{C}$ | | 0.702 |
| MTPT CSMA[50] | 224x224 | $\mathcal{M}_\mathcal{I}$ | $\mathcal{M}_\mathcal{T}$ | $\mathcal{C}$ | | 0.688 |
| MMQ[15] | 84x84 | $\mathcal{M}_\mathcal{I}$ | $\mathcal{N}_\mathcal{T}$ | $\mathcal{C}$ | | 0.670 |
| MEVF[51] | - | $\mathcal{N}/\mathcal{A}$ | $\mathcal{N}_\mathcal{T}$ | $\mathcal{C}$ | | 0.627 |
| BioMedBLIP[3] | 224x224 | $\mathcal{M}_\mathcal{I}$ | $\mathcal{M}_\mathcal{I}$ | $\mathcal{B}$ | | 0.357 |
| MedFlamingo (Few shots) [60] | 224x224 | $\mathcal{M}_\mathcal{I}$ | $\mathcal{M}_\mathcal{T}$ | $\mathcal{G}$ | | 0.200 |
| PubMedClip [56](MEVF) (QCR) | 224x224 | $\mathcal{M}_\mathcal{I}$ | $\mathcal{M}_\mathcal{T}$ | $\mathcal{C}$ | | 0.665 0.721 |
| HQS-VQA[42] | 224x224 | $\mathcal{N}_\mathcal{I}$ | $\mathcal{N}_\mathcal{T}$ | $\mathcal{C}$ | 0.411 | |
| **VQA-Med 2019** | | | | | | |
| MedfuseNet[19] | 224x224 | $\mathcal{N}_\mathcal{I}$ | $\mathcal{N}_\mathcal{T}$ | $\mathcal{B}$ | 0.27 | 0.789 |
| MMBERT[23] | 224x224 | $\mathcal{N}_\mathcal{I}$ | $\mathcal{N}_\mathcal{T}$ | $\mathcal{G}$ | 0.690 | 0.672 |
| VB-MVQA[74] | 224x224 | $\mathcal{N}_\mathcal{I}$ | $\mathcal{N}_\mathcal{T}$ | $\mathcal{B}$ | 0.673 | 0.668 |
| QFNS[75] | 224x224 | $\mathcal{N}_\mathcal{I}$ | $\mathcal{N}_\mathcal{T}$ | $\mathcal{B}$ | 0.638 | 0.657 |
| MHKD-MVQA[73] | - | $\mathcal{N}_\mathcal{I}$ | $\mathcal{N}_\mathcal{T}$ | $\mathcal{G}$ | 0.667 | 0.650 |
| WADAN (NP) [74] | - | $\mathcal{N}_\mathcal{I}$ | $\mathcal{N}_\mathcal{T}$ | $\mathcal{B}$ | 0.662 | 0.645 |
| CGMVQA[44] | - | $\mathcal{N}_\mathcal{I}$ | $\mathcal{N}_\mathcal{T}$ | $\mathcal{G}$ | 0.659 | 0.640 |
| SupCon-SB[70] | 224x224 | $\mathcal{N}_\mathcal{I}$ | $\mathcal{N}_\mathcal{T}$ | $\mathcal{C}$ | 0.643 | 0.628 |
| Hanlin[28] | 224x224 | $\mathcal{N}_\mathcal{I}$ | $\mathcal{N}_\mathcal{T}$ | $\mathcal{C}$ | 0.644 | 0.620 |
| Minhvu[45] | 448x448 | $\mathcal{N}_\mathcal{I}$ | $\mathcal{N}_\mathcal{T}$ | $\mathcal{C}$ | 0.634 | 0.616 |
| TUA1[54] | - | $\mathcal{N}_\mathcal{I}$ | $\mathcal{N}_\mathcal{T}$ | $\mathcal{B}$ | 0.633 | 0.60 |
| MMS | - | $\mathcal{N}_\mathcal{I}$ | $\mathcal{N}_\mathcal{T}$ | $\mathcal{C}$ | 0.593 | 0.566 |
| JUST[29] | - | $\mathcal{N}_\mathcal{I}$ | $\mathcal{N}/\mathcal{A}$ | $\mathcal{B}$ | 0.591 | 0.534 |
| Team_PwC_Med[76] | 512x512 | $\mathcal{N}_\mathcal{I}$ | $\mathcal{N}_\mathcal{T}$ | $\mathcal{B}$ | 0.534 | 0.488 |
| Techno[30] | - | $\mathcal{N}/\mathcal{A}$ | $\mathcal{N}/\mathcal{A}$ | $\mathcal{C}$ | 0.486 | 0.462 |
| Abhishek-thanki[31] | - | $\mathcal{N}_{\mathcal{I}}$ | $\mathcal{N}_{\mathcal{T}}$ | $\mathcal{G}$ | 0.462 | 0.160 |
| BPI-MVQA[69] | 224x224 | $\mathcal{N}_{\mathcal{I}}$ | $\mathcal{M}_{\mathcal{T}}$ | $\mathcal{B}$ | 0.687 | |
| **VQA-Med 2020** | | | | | | |
| AIML[20] | - | $\mathcal{N}/\mathcal{A}$ | $\mathcal{N}/\mathcal{A}$ | $\mathcal{C}$ | 0.542 | 0.496 |
| TheInception_ Team[32] | - | $\mathcal{N}/\mathcal{A}$ | $\mathcal{N}/\mathcal{A}$ | $\mathcal{C}$ | 0.511 | 0.480 |
| bumjun_jung[33] | 224x224 | $\mathcal{N}_{\mathcal{I}}$ | $\mathcal{M}_{\mathcal{T}}$ | $\mathcal{C}$ | 0.502 | 0.466 |
| HCP-MIC[47] | - | $\mathcal{N}/\mathcal{A}$ | $\mathcal{M}_{\mathcal{T}}$ | $\mathcal{C}$ | 0.462 | 0.426 |
| NLM[43] | - | $\mathcal{N}_{\mathcal{I}}$ | $\mathcal{N}_{\mathcal{T}}$ | $\mathcal{C}$ | 0.441 | 0.400 |
| HARENDRA-KV[34] | 512X512 | $\mathcal{N}/\mathcal{A}$ | $\mathcal{N}_{\mathcal{T}}$ | $\mathcal{G}$ | 0.439 | 0.376 |
| Shengyan[35] | - | $\mathcal{N}_{\mathcal{I}}$ | $\mathcal{N}/\mathcal{A}$ | $\mathcal{G}$ | 0.412 | 0.376 |
| kedvqa[36] | 255x255 | $\mathcal{N}/\mathcal{A}$ | $\mathcal{N}_{\mathcal{T}}$ | $\mathcal{C}$ | 0.35 | 0.314 |
| **VQA-Med 2021** | | | | | | |
| SYSU_HCP[22] | 244x224 | $\mathcal{N}_{\mathcal{I}}$ | $\mathcal{N}/\mathcal{A}$ | $\mathcal{C}$ | 0.416 | 0.382 |
| Yunnan_University[37] | 224x224 | $\mathcal{N}_{\mathcal{I}}$ | $\mathcal{M}_{\mathcal{T}}$ | $\mathcal{C}$ | 0.402 | 0.362 |
| Lijie[38] | 224x224 | $\mathcal{N}_{\mathcal{I}}$ | $\mathcal{M}_{\mathcal{T}}$ | $\mathcal{C}$ | 0.352 | 0.316 |
| TAM[49] | 128x 128 | $\mathcal{N}/\mathcal{A}$ | $\mathcal{N}_{\mathcal{T}}$ | $\mathcal{G}$ | 0.255 | 0.222 |
| Sheerin[53] | 224x224 | $\mathcal{N}_{\mathcal{I}}$ | $\mathcal{N}/\mathcal{A}$ | $\mathcal{G}$ | 0.227 | 0.196 |
| **SLAKE** | | | | | | |
| CMMO[21] | 384x384 | $\mathcal{M}_{\mathcal{I}}$ | $\mathcal{M}_{\mathcal{T}}$ | $\mathcal{G}$ | | 0.872 |
| LlaVA Med[55] | | $\mathcal{M}_{\mathcal{I}}$ | $\mathcal{M}_{\mathcal{T}}$ | $\mathcal{B}$ | | 0.853 |
| CPRD[50] | 224x224 | $\mathcal{N}_{\mathcal{I}}$ | $\mathcal{N}_{\mathcal{T}}$ | $\mathcal{C}$ | | 0.821 |
| MISS[63] | 480x480 | $\mathcal{N}_{\mathcal{I}}$ | $\mathcal{N}_{\mathcal{T}}$ | $\mathcal{G}$ | | 0.820 |
| MITER[64] | 448x448 | $\mathcal{M}_{\mathcal{I}}$ | $\mathcal{M}_{\mathcal{T}}$ | $\mathcal{C}$ | | 0.812 |
| VQA-Adapter[66] | 224x224 | $\mathcal{N}_{\mathcal{I}}$ | $\mathcal{N}_{\mathcal{T}}$ | $\mathcal{C}$ | | 0.810 |
| BioMedBLIP[3] | 224x224 | $\mathcal{M}_{\mathcal{I}}$ | $\mathcal{M}_{\mathcal{T}}$ | $\mathcal{B}$ | | 0.808 |
| VB-MVQA[74] | 224x224 | $\mathcal{N}_{\mathcal{I}}$ | $\mathcal{N}_{\mathcal{T}}$ | $\mathcal{B}$ | 0.786 | 0.787 |
| M2I2[62] | 384x384 | $\mathcal{N}_{\mathcal{I}}$ | $\mathcal{N}_{\mathcal{T}}$ | $\mathcal{G}$ | | 0.768 |
| PubMedClip [56]<br>(MEVF)<br>(QCR) | 224x224 | $\mathcal{M}_{\mathcal{I}}$ | $\mathcal{M}_{\mathcal{T}}$ | $\mathcal{G}$ | | 0.780<br>0.801 |
| **OVQA** | | | | | | |
| MMBERT[23] | 224x224 | $\mathcal{N}_{\mathcal{I}}$ | $\mathcal{N}_{\mathcal{T}}$ | $\mathcal{G}$ | | 0.633 |

**Supplementary C**

This appendix provides a comprehensive overview of the survey conducted as part of this study. The survey was administered via Google Forms and consisted of three pages, with all questions presented in a multiple-choice format. A total of 50 responses were collected from participants across India (n=40) and Thailand (n=10). Link to survey form: Link here

**First Page: Introduction and Consent**

The first page of the survey provided participants with a brief introduction to the **Medical Visual Question Answering (MedVQA)** system, outlining its purpose and potential applications in radiology. This section also included a consent form to ensure that participants understood the study objectives and voluntarily agreed to participate.

**1. Would you be willing to participate in this survey?**

a. Yes
b. No

**Second Page: Understanding Participants' Background and Expertise**

This section aimed to gather demographic and professional background information, including participants' specialization, years of experience, and familiarity with artificial intelligence (AI) in medical imaging. The purpose of these questions was to contextualize responses and assess the level of expertise among participants regarding AI applications in healthcare. The questions, along with their qualitative analysis, are presented below.

**Questions and their qualitative analysis:**
**1. Which country you belong to?**
a. India
b. Thailand

**2. What is your primary area of specialization in healthcare?**
a. Physician
b. Surgeon
c. Radiologist
d. Clinicians

**3. How many years of experience do you have as a healthcare professional?**
a. Less than 5 years
b. 5–10 years
c. 11–20 years

d. More than 20 years

**4. How familiar are you with artificial intelligence in medical imaging?**
(Rate your familiarity on a scale of 1 to 5, where 1 = Not familiar and 5 = Very familiar)
a. 1: Not familiar
b. 2: Slightly familiar
c. 3: Moderately familiar
d. 4: Quite familiar
e. 5: Very familiar

**5. What is your current level of involvement with AI technologies in your workflow?**
a. 1 (None)
b. 2 (Occasional use, e.g., automated tools for basic tasks)
c. 3 (Moderate use, e.g., some integration into workflow)
d. 4 (Frequent use, e.g., advanced AI tools regularly integrated)
e. 5 (Expert use, e.g., actively contributing to or managing AI implementation)

**Third Page: Perspectives on MedVQA Integration**

This section explores participants' perspectives on the potential integration of MedVQA into radiology workflows. It seeks to gather insights regarding the applicability and relevance of existing MedVQA models, as well as their perceived utility in clinical practice. The questions, along with the qualitative analysis of the responses, are presented below.

**Questions and their qualitative analysis:**

**1. How do you perceive the potential role of MedVQA systems in supporting your daily healthcare practice?**
a. 1 (Not useful at all)
b. 2 (Somewhat useful)
c. 3 (Extremely useful)

**2. Which aspects of MedVQA models do you find most relevant to your workflow?**
a. Image interpretation (e.g., identifying abnormalities)
b. Generating reports based on images
c. Providing diagnostic suggestions
d. Integrating with existing radiology systems (e.g., PACS, RIS)
e. Assisting in time-sensitive cases (e.g., emergency scans)
f. Providing interpretability and transparency of results

**3. In your opinion, how important is explainability or the ability for MedVQA models to provide understandable reasoning behind their answers (e.g., "Why was this diagnosis suggested?")?**
a. 1 (Not important at all)
b. 2 (Slightly important)
c. 3 (Moderately important)
d. 4 (Very important)
e. 5 (Extremely important)

4. How do you think MedVQA models can improve the efficiency of your work?

a. By speeding up the image interpretation process
b. By reducing manual effort in generating reports
c. By assisting in prioritizing urgent cases
d. By decreasing cognitive load (e.g., reducing decision fatigue)

**5. What would be the most important feature you would look for in a MedVQA system before adopting it into your practice?**
a. High accuracy of answers
b. Fast processing time
c. Easy integration with existing systems
d. User-friendly interface
e. Detailed explanations for diagnoses
f. Continuous learning and adaptation to new cases

**6. Do you believe MedVQA systems can significantly improve patient outcomes in your practice?**
a. 1 (No improvement)
b. 2 (Minimal improvement)
c. 3 (Moderate improvement)
d. 4 (Substantial improvement)
e. 5 (Significant improvement)

**7. Do questions about modality, size, color, or the presence of an organ (e.g., "What is the primary abnormality in the image? Is it a fracture?" or "What is the size of the tumor?") fall under educational relevance, clinical relevance, or both in MedVQA systems?**
a. Educational relevance
b. Clinical relevance
c. Both

**8. Would you categorize yes/no questions (e.g., "Is this a fracture?" or "Is there evidence of malignancy?") in MedVQA systems as primarily educational, clinically relevant, or more suited for a dialogue-based interaction?**
a. Dialogue-based interaction
b. Educational Relevance
c. Clinical Relevance

**9. How do you perceive the integration of external knowledge such as Electronic Health Records (EHR) or domain-specific knowledge into MedVQA systems?**
a. Very beneficial
b. Somewhat beneficial
c. Not beneficial
d. Unsure

**10. If you think it's beneficial, which type of external knowledge would you prefer to integrate?**
a. EHR data
b. Domain-specific knowledge (e.g., expert radiology knowledge)
c. Medical literature or guidelines
d. All
e. Not sure

**11. What type of MedVQA system do you think would be more useful in your practice?**
a. A system that answers a specific question about the image (e.g., "Is there a fracture in this image?")
b. A system that interacts in a dialogue form (like a chatbot that asks and answers multiple questions in sequence)
c. Both types

**12. Would you trust a MedVQA system that is trained using manually curated datasets with the help of doctors, or would you trust one trained with synthetic data (generated by AI models)?**
a. I trust systems using manually curated datasets with the help of doctors
b. I trust systems using synthetic data
c. I trust both equally
d. I trust neither
e. Unsure

**13. What is your opinion on MedVQA systems that use single images versus multiple views (e.g., X-ray, CT scans) to answer questions?**
a. I prefer single images for answering question
b. I prefer multiple views for more accurate answers
c. I think both are equally useful
d. Unsure

**14. Which type of MedVQA system would you prefer?**
a. 1 (Strongly prefer a system specialized for one anatomy)
b. 2 (Somewhat prefer a system specialized for one anatomy)
c. 3 (No preference)
d. 4 (Somewhat prefer a general model)
e. 5 (Strongly prefer a general model)